\documentclass[runningheads]{llncs}
\usepackage{graphicx}
\usepackage{amsmath,amssymb} % define this before the line numbering.
\usepackage{color}
\usepackage[width=122mm,left=12mm,paperwidth=146mm,height=193mm,top=12mm,paperheight=217mm]{geometry}

\usepackage{epsfig}
\usepackage{graphicx} 
\usepackage{subfig}
\usepackage{setspace}
\usepackage{mathtools} 
\usepackage{algorithmicx}
\usepackage[linesnumbered, ruled]{algorithm2e}
\usepackage{algpseudocode}
\usepackage{multirow}
\usepackage{color, colortbl}
\definecolor{LightGray}{rgb}{0.9,0.9,0.9}
\usepackage{tabularx}
\usepackage{url}

\newcommand{\etal}{\mbox{\emph{et al. }}}

\begin{document}
%\linenumbers
\pagestyle{headings}
\mainmatter
\def\ECCV16SubNumber{}  % Insert your submission number here

\title{Monocular Visual Odometry  with a Rolling Shutter Camera} % Replace with your title

\titlerunning{}

\authorrunning{}

\author{Chang-Ryeol Lee and Kuk-Jin Yoon}

\institute{School of Electrical Engineering and Computer Science,\\
	Gwanju Institue Science and Technology\\
	\email{ \{crlee,kjyoon\}@gist.ac.kr}
}

\maketitle

\begin{abstract}

Rolling Shutter (RS) cameras have become popularized because of low-cost imaging capability.
However, the RS cameras suffer from undesirable artifacts when the camera or the subject is moving, or illumination condition changes. 
For that reason, Monocular Visual Odometry (MVO) with RS cameras produces inaccurate ego-motion estimates. 
Previous works solve this RS distortion problem with motion prediction from images and/or inertial sensors. 
However, the MVO still has trouble in handling the RS distortion when the camera motion changes abruptly (\textit{e.g.} vibration of mobile cameras causes extremely fast motion instantaneously). 
To address the problem, we propose the novel MVO algorithm in consideration of the geometric characteristics of RS cameras.
The key idea of the proposed algorithm is the new RS essential matrix which incorporates the instantaneous angular and linear velocities at each frame. 
Our algorithm produces accurate and robust ego-motion estimates in an online manner, and is applicable to various mobile applications with RS cameras.
The superiority of the proposed algorithm is validated through quantitative and qualitative comparison on both synthetic and real dataset.

%The abstract should summarize the contents of the paper. LNCS guidelines
%indicate it should be at least 70 and at most 150 words. It should be set in 9-point
%font size and should be inset 1.0~cm from the right and left margins.
%\dots
\end{abstract}

\section{Introduction} \label{sec:intro}

% monocular rolling shutter visual odometry의 중요성/필요성
Odometry that estimates 6-DOF ego-motion is a crucial technology for mobile applications and robotics applications.
Visual Odometry (VO) using cameras has been extensively studied for robot navigation \cite{Moravec:IJCAI:1979} and autonomous driving \cite{Geiger:CVPR:2012} for decades. 
Practically, VO has distinct advantages in GPS-denied environments such as urban, military, underwater and indoor areas, and provides less drifted results compared to Wheel Odometry (WO) and Inertial Odometry (IO).
Especially, Monocular Visual Odometry (MVO) has been actively studied for a decade because of its compactness and price competitiveness \cite{Davison:PAMI:2007}.

\begin{figure}[tb]
	\centering	
	\includegraphics[width=0.99\linewidth]{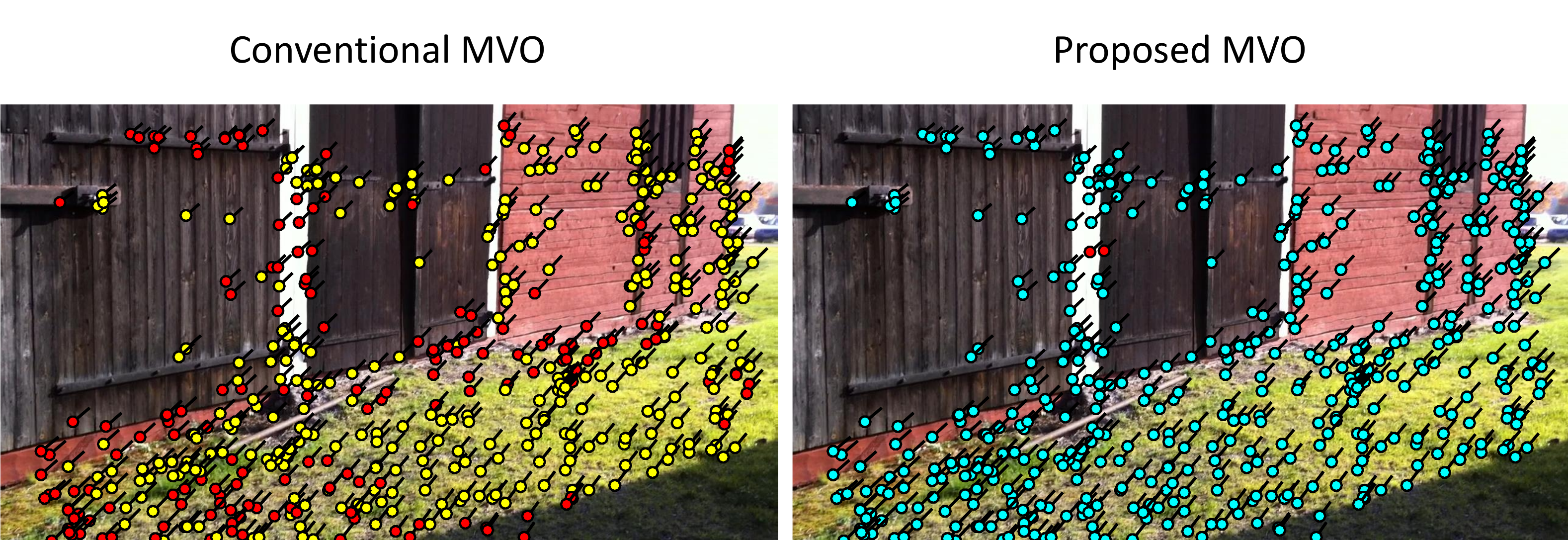}
	\caption{Results of the conventional MVO and the proposed MVO. In both figures, dots represent tracked feature points and lines represent their displacements. (Left) yellow and red dots represent the inliers and outliers of the conventional MVO, respectively. (Right) cyan and red dots represent the inliers and outliers of the proposed MVO, respectively. We select 500 strongest points, and the numbers of inliers of the proposed MVO and the conventional MVO are 498 and 324, respectively.}		
	\label{fig:intro}
	\vspace{-4mm}
\end{figure}

For the MVO, Rolling Shutter (RS) cameras, which capture the image line-by-line, are more preferable than Global Shutter (GS) cameras, which capture all image lines at once, because of the low-cost imaging capability.
However, the MVO with an RS camera becomes a challenging problem when the camera (or subject) is moving or illumination condition changes, because the all lines of an RS image are obtained from different poses.
The pose changing during the RS image capturing process causes unmodeled noises and outliers in feature points-based ego-motion estimation.
To correct the RS artifacts, researchers exploit the predicted motion information from temporally neighboring frames with the assumption on smooth camera motion \cite{Klein:ISMAR:2009} \cite{Hedborg:CVPR:2012} and/or inertial sensors \cite{Karpenko:Tech:2011} \cite{Jia:MMSP:2012}.
However, it is difficult to predict the RS artifacts when the camera motion changes abruptly as in hand-held and/or vehicle-attached cameras
(e.g. vibration of the car due to the uneven ground plane),
because the camera motion changes extremely fast instantaneously.  
This unpredictable RS artifact dramatically reduces the number of inliers in ego-motion estimation and results in inaccurate and inconsistent ego-motion estimation.
Figure \ref{fig:intro} shows a distorted RS image and the effect of RS artifacts on ego-motion estimation (left).
%
%The proposed algorithm provides a larger number of inliers than conventional MVO and this leads to more accurate ego-motion estimation.

% our solution
In this paper, we propose a novel MVO algorithm in consideration of the geometric characteristics of RS cameras. In the proposed algorithm, we jointly estimate the relative camera transformation between two frames and instantaneous camera motion, which consists of the linear and the angular velocities, at each frame. 
The key idea of the proposed MVO algorithm is the new RS essential matrix which incorporates the instantaneous angular and linear velocities at each frame.
The RS essential matrix is highly nonlinear and has 17-DOF (relative rotation/translation and RS linear and angular velocities), and we adopt the Levenberg-Marquardt algorithm to estimate relative camera transformation and instantaneous camera motion from the RS essential matrix. By using the RS essential matrix, the proposed algorithm provides a larger number of inliers than conventional MVO as shown in Fig.~\ref{fig:intro}. % and this leads to more accurate ego-motion estimation.
Consequently, our algorithm produces accurate and robust ego-motion estimates in an online manner, and it is applicable to various mobile applications. 
The proposed approach can also handle RS artifacts caused by smooth and/or abrupt motion.
Besides, our work can be exploited to provide an initial solution for time-delayed/off-line MVO algorithms.
The contributions of this paper can be summarized as follows.

$\bullet$   Introduction of the instantaneous camera motion model for the RS camera problem,

$\bullet$   Formulation of the RS essential matrix,

$\bullet$	Proposition of the joint estimation algorithm of relative pose and instantaneous motion with only two images.

This paper is organized as follows.
We review related works in Sec. \ref{sec:related_work}.
Then, we define the terminologies to make formulation and explanation clear in Sec. \ref{sec:term}.
The RS camera geometry is then explained in Sec. \ref{sec:rs_geometry}. 
We describe the RS essential matrix which incorporates RS camera geometry in Sec. \ref{sec:RS_essential}, and explain how to estimate the relative rotation and translation from the RS essential matrix in Sec. \ref{sec:motion_estimation}.
We show experimental results in Sec. \ref{sec:experiments}. Finally, we discuss the limitation and future works, and conclude the paper in Sec. \ref{sec:conclusion}.

\vspace{-1mm}
\section{Related Works} \label{sec:related_work}

% rolling shutter effect handling for various application
The RS effect has been dealt with in several computer vision problems, such as Perspective-n-Point (PnP) problem that requires 3D point clouds generated by a GS camera.
Ait-Aider \etal estimated the pose and velocity of fast moving objects in a single image with a rolling shutter camera \cite{Ait:ECCV:2006}.
The nonlinear and linear models were proposed for general and planar objects, respectively.
Magerand \etal extended this work with a polynomial rolling shutter model and the constrained global optimization \cite{Henrion:OMS:2009} \cite{Magerand:ECCV:2012}.
Albl \etal proposed a double linearized rolling shutter for an efficient estimation \cite{Albl:CVPR:2015}.
This work remarkably increased the accuracy of motion estimation and the number of inliers.

% rolling shutter effect handling for motion estimation problem with inertial sensor
On the other hand, inertial sensors are powerful options when estimating ego-motion with an RS camera.
Karpenko \etal exploited gyroscopes to correct RS effects and stabilized video on a smart-phone \cite{Karpenko:Tech:2011}.
Synchronization between a gyroscope and a camera was performed by comparing angular velocity measurements obtained from the gyroscope and feature displacements computed from the video.
Jia and Evans estimated camera orientation on the Bayesian estimation framework with inertial measurements, and also corrected RS effects of images \cite{Jia:MMSP:2012}.
Guo \etal proposed the ego-motion estimation framework by fusing an inertial sensor and a camera while considering artifacts such as RS effects and synchronization between inertial measurements and images occurred in mobile devices \cite{Guo:RSS:2014}.
%

% rolling shutter effect handling for motion estimation problem
Handling RS distortion for ego-motion estimation only with a monocular RS camera (without using inertial sensors) has been also studied.
Klein and Murray corrected the RS effect by using the camera velocity estimated in the ego-motion estimation framework \cite{Klein:ISMAR:2009}.
Since this work focused on the efficient of an algorithm, a simple strategy was selected.
Hedborg \etal proposed to use temporal information from an image sequence \cite{Hedborg:ICCVW:2011} \cite{Hedborg:CVPR:2012}.
The proposed RS bundle adjustment is a powerful scheme, but it is time-consuming and an off-line algorithm. 
Saurer \etal handled the RS distortion for the 3D reconstruction problem because the RS effects and the 3D structure of scenes are highly related to each other~\cite{Saurer:ICCV:2013}.

Our work is also based only on a monocular RS camera and estimates ego-motion in consideration of the RS effects in an online manner (\textit{i.e.} two-frame relative motion estimation).
Consequently, unlike previous works, the proposed algorithm does not require any of 3D point clouds, additional inertial sensors, and temporal image information (\textit{i.e.} video).

\begin{figure}[tb]
	\centering	
	\includegraphics[width=0.80\linewidth]{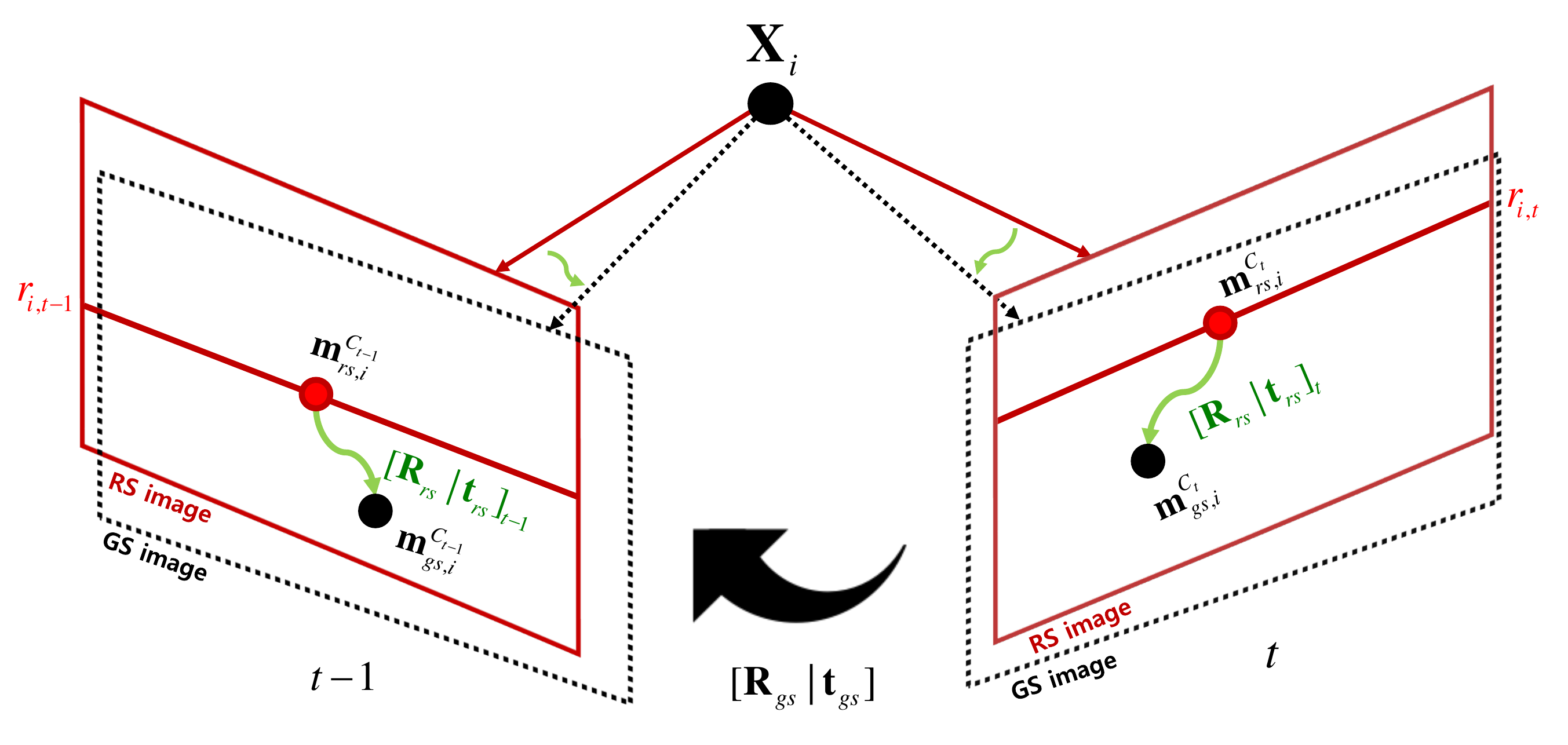}	
	\caption{Two-view geometry of an RS camera.  GS transformation is shown with a black arrow and  RS transformations are shown with green arrows. GS and RS images are represented as red solid lines and black dashed lines, respectively. }		
	\label{fig:two_view_geometry} 
	\vspace{-2mm}
\end{figure}

\vspace{-2mm}
\section{Terminology} \label{sec:term}

Before presenting the RS camera geometry and the proposed algorithm, we define several terms on the two-view geometry of an RS camera in Fig. \ref{fig:two_view_geometry}. % to make our formulation clear.
%
%The GS image is the image that the RS distortion is removed from the original RS image.
The RS image is the image containing the RS distortion (a GS image $+$ the RS distortion), and the subscript $gs$ and $rs$ indicate a global shutter camera and a rolling shutter camera, respectively. 
The transformation between two RS images is composed of one GS transformation and two RS transformation defined as follows.

\noindent
\textbf{1)}~{\textit{GS~transformation}}:  rotation $\mathbf{R}_{gs} \in SO(3)$ and translation $\mathbf{t}_{gs} \in \mathbb{R}^{3}$ between the two undistorted GS images from $t$ frame to $t-1$ frame. 

\noindent
\textbf{2)}~{\textit{RS~transformation}}:  row-wise rotation $\mathbf{R}_{rs}(r_{i}) \in SO(3)$ and translation $\mathbf{t}_{rs}(r_{i}) \in \mathbb{R}^{3}$ from the distorted RS image to the undistorted GS image at   frame $t$ (or $t-1$) (refer Fig.~\ref{fig:two_view_geometry}), where $r_{i}$ denotes the row of a feature point, the subscript $i$ is the index of a feature point. The time range of the RS transformation in an image is from $0$ to $ \left(n_{row}-1\right) \cdot \tau$, where $n_{row}$ is the number of rows and $\tau$ is the exposure time for one line.

\noindent
\textbf{3)}~{\textit{Instantaneous~RS~motion}}:  angular and linear velocities $\mathbf{w} \in \mathbb{R}^{3}$, $\mathbf{v} \in \mathbb{R}^{3}$ around $r_i=0$. 

%Consequently, the transformation between two RS images is composed of one GS~transformation and two RS~transformation.
%
%The subscript $gs$ and $rs$ indicate a global shutter camera and a rolling shutter camera.

\vspace{-2mm}
\section{Rolling Shutter Camera Geometry} \label{sec:rs_geometry}

In general GS camera geometry, a 3D point $\mathbf{X}_i^{W} \in \mathbb{P}^3 $ in the homogeneous world coordinate is projected to a 2D point $\mathbf{m}_i \in \mathbb{R}^2$ in the image coordinate with camera intrinsic parameters (focal length, skew, position of a principal point) and extrinsic parameters (rotation and translation). Here, superscript $W$ indicates the world coordinate.
This projection can be described as
\begin{equation}
	\mathbf{m}_{gs,i} = \left[ \begin{matrix}  c_i \\  r_i  \end{matrix} \right]_{gs} \sim  \mathbf{K} \left[ \ \mathbf{R} \ | \ \mathbf{t} \ \right] \mathbf{X}_i^{W}, 
\end{equation}
where $\mathbf{K} \in \mathbb{R}^{3 \times 3}$ is an intrinsic matrix,  $\mathbf{R} \in SO(3)$ is a rotation matrix, and  $\mathbf{t} \in \mathbb{R}^{3}$ is a translation vector with respect to the world coordinate.
The RS camera geometry is derived from this well-known GS camera geometry.

In an RS camera, a 3D point $\mathbf{X}_i^{W} \in \mathbb{P}^{3}$ is projected to the 2D point $\mathbf{m}_{rs,i} \in \mathbb{R}^{2}$
with the line-by-line exposure, that is, the RS transformation as
\begin{equation}
\mathbf{m}_{rs,i} = \left[ \begin{matrix}  c_i \\  r_i \end{matrix} \right]_{rs} \sim 
\mathbf{K} 
\left[
\begin{matrix}
\ \mathbf{R}_{rs}(r_i) & \mathbf{t}_{rs}(r_i) \
\end{matrix}
\right]
\left[
\begin{matrix}
	\mathbf{R} & \mathbf{t} \\
	\mathbf{0} & 1
\end{matrix}
\right]
\mathbf{X}_i^{W}.
\end{equation}
The RS transformation $[\mathbf{R}_{rs}(r_i)$, $\mathbf{t}_{rs}(r_i)]$ is approximated as linear functions of  the image row  $r_{i}$ with instantaneous RS motion $\mathbf{w},\mathbf{v}$ as 
\begin{equation}
\mathbf{R}_{rs}(r_i) \simeq \mathbf{R}_{rs}(r_i\mathbf{w})  =  \mathbf{I}_{3} + r_i \tau \left[ \begin{matrix}   0  & -w_z & w_y  \\    w_z &  0  & -w_x  \\  -w_y & w_x & 0  \end{matrix} \right],  
\end{equation}
\begin{equation}
\  \  \  \ \mathbf{t}_{rs}(r_i) \simeq \mathbf{t}_{rs}(r_i\mathbf{v}) = r_i \tau \left[ \begin{matrix}  v_x \\  v_y \\ v_z \end{matrix} \right] , 
\end{equation}
where $\tau$ is an exposure time for one line of an RS camera.

The relation between the GS image and the  RS image can be expressed as 
\begin{equation}
\mathbf{m}_{rs,i} 
\sim 
\mathbf{K} 
\left[
\begin{matrix}
\ \mathbf{R}_{rs}(r_i\mathbf{w}) & \mathbf{t}_{rs}(r_i\mathbf{v}) \
\end{matrix}
\right]
\left[ 
\
\begin{matrix}
\mathbf{H}^{-1} & \mathbf{0}\\
\mathbf{0}  & 1
\end{matrix} 
\
\right]
\left[ 
\
\begin{matrix}
\mathbf{K}^{-1} & \mathbf{0}\\
\mathbf{0}  & 1
\end{matrix} 
\
\right]
\left[ 
\
\begin{matrix}
\tilde{\mathbf{m}}_{gs,i} \\
\mathbf{1} 
\end{matrix} 
\
\right],
\end{equation}
where $\mathbf{H} \in \mathbb{R}^{3 \times 3}$ is a back-projection matrix and 
$\tilde{\mathbf{m}}_i \in \mathbb{R}^3$ is the position of a feature point in the normalized image coordinates (i.e. $ \tilde{\mathbf{m}}_i = \left[ \ c_i, \ r_i, \ 1 \ \right]^T$ ).
Green arrows in Fig. \ref{fig:two_view_geometry} indicate the transformation between feature correspondences in the GS and the RS images.

\section{Rolling Shutter Essential Matrix} \label{sec:RS_essential}

To estimate ego-motion in an online manner, we focus on the relative transformation between two consecutive frames.
Efficient MVO can be achieved up to scale by concatenating the relative transformation estimates \cite{Geiger:IV:2011}.
Estimating relative transformation with a GS camera is well-formulated in the fundamental/essential matrix estimation problem\cite{Hartley:BOOK:2003} as
\begin{equation}
\left( \tilde{\mathbf{m}}_{gs}^{C_{t-1}} \right)^{T}\mathbf{F}_{gs}\tilde{\mathbf{m}}_{gs}^{C_{t}} = 0 , \ \ \mathbf{F}_{gs} = \mathbf{K}^{-T}  \mathbf{E}_{gs} \mathbf{K}^{-1}, \ \ \mathbf{E}_{gs} = {\left\lfloor \mathbf{t}_{gs} \right\rfloor_{\times} \mathbf{R}_{gs}},
\end{equation}
where superscript $C$ indicates the camera coordinate, the subscript $t-1$ and  $t$  indicate time indices of two consecutive frames, and
$\mathbf{F}_{gs} \in \mathbb{R}^{3 \times 3}$ is a GS fundamental matrix.

The 2D feature points $\tilde{\mathbf{m}}_{rs}^{C_{t-1}}$, $\tilde{\mathbf{m}}_{rs}^{C_{t}}$ in two consecutive RS images  and RS fundamental matrix $\mathbf{F}_{rs}$  satisfy the following constraint:
\begin{equation}
\left(\tilde{\mathbf{m}}_{rs}^{C_{t-1}}\right)^{T} \mathbf{F}_{rs} \tilde{\mathbf{m}}_{rs}^{C_{t}} = 0, \ \ \mathbf{F}_{rs} = \mathbf{K}^{-T}  \mathbf{E}_{rs} \mathbf{K}^{-1}, \ \ \left(\tilde{\mathbf{x}}_{rs}^{C_{t-1}}\right)^{T} \mathbf{E}_{rs} \tilde{\mathbf{x}}_{rs}^{C_{t}} = 0,
\end{equation}
where $\tilde{\mathbf{x}} \in \mathbb{P}^{2}$ is  a feature point in the homogeneous camera coordinate (i.e. $ \tilde{\mathbf{x}} = \left[ \ x, \ y, \ 1 \ \right]^T  = \mathbf{K}^{-1} \tilde{\mathbf{m}}$ )

Now, we derive the essential matrix $\mathbf{E}_{rs}$ which incorporates instantaneous RS motion at each frame, that is, the RS essential matrix. 
With the given scale of feature points, the RS essential matrix satisfy following constraint:
\begin{equation}
\left(\mathbf{x}_{rs}^{C_{t-1}}\right)^{T} \mathbf{E}_{rs} \mathbf{x}_{rs}^{C_{t}} = 0,
\label{eq:rs_essential_matrix}
\end{equation}
where $\mathbf{x}^{C_t} = \mathbf{H} \tilde{\mathbf{x}}^{C_t} \in \mathbb{R}^{3}$ is a feature point in the camera coordinate.
The feature points $\mathbf{x}_{rs}^{C_{t-1}}$ and  $\mathbf{x}_{rs}^{C_{t}}$ in the RS camera coordinates are expressed with respect to the feature points $\mathbf{x}_{gs}^{C_{t}}$ in the GS camera coordinate. For brevity, we denote $\mathbf{R}_{rs}(r_i \mathbf{w}^{C_t})$ and $\mathbf{t}_{rs}(r_i \mathbf{v}^{C_t})$ as $\mathbf{R}_{rs}^{C_t}$ and $\mathbf{t}_{rs}^{C_t}$. Then, 
\begin{equation}
\small
\begin{aligned}
\mathbf{x}_{rs}^{C_{t-1}} 
= 
\left[
\begin{matrix}
\ \mathbf{R}_{rs}^{C_{t-1}}  &   \mathbf{t}_{rs}^{C_{t-1}} \ 
\end{matrix}
\right]^{-1}
\left[
\begin{matrix}
\mathbf{R}_{gs}  & \mathbf{t}_{gs}  \\
\mathbf{0}  & 1
\end{matrix}
\right]^{-1}
\mathbf{X}_{gs}^{C_{t}} \ \
s.t. \ \ \mathbf{X}_{gs}^{C_{t}} 
=
\left[
\begin{matrix}
\mathbf{R}_{gs}  & \mathbf{t}_{gs}  \\
\mathbf{0}  & 1
\end{matrix}
\right]
\mathbf{X}_{gs}^{C_{t-1}},
\end{aligned}
\label{eq:two_frame_relation_1}
\end{equation} 
\begin{equation}
\mathbf{x}_{rs}^{C_{t}} 
=
\left[
\begin{matrix}
\ \mathbf{R}_{rs}^{C_{t}}  & \mathbf{t}_{rs}^{C_{t}} \ \ 
\end{matrix}
\right]^{-1}
\mathbf{X}_{gs}^{C_{t}},
\label{eq:two_frame_relation_2}
\end{equation} 
where $\mathbf{X}^{C} \in \mathbb{P}^{3}$ is the feature point in the homogeneous camera coordinate.

The 3D feature points in RS two consecutive frames are converted to 3D points in the GS camera with Eq.~(\ref{eq:two_frame_relation_1}) and Eq.~(\ref{eq:two_frame_relation_2}).
Thus, the constraint Eq.~(\ref{eq:rs_essential_matrix}) is converted to 
\begin{equation}
\small
{ 
\left( 
\left[
\begin{matrix}
\ \mathbf{R}_{rs}^{C_{t-1}}  &   \mathbf{t}_{rs}^{C_{t-1}} 
\end{matrix}
\right]^{-1}
\left[
\begin{matrix}
\mathbf{R}_{gs}  & \mathbf{t}_{gs}  \\
\mathbf{0}  & 1
\end{matrix}
\right]^{-1}
\mathbf{X}_{gs}^{C_{t}}
\right)}^{T} 
\mathbf{E}_{rs} 
{ \left(  
\left[
\begin{matrix}
\ \mathbf{R}_{rs}^{C_{t}}  & \mathbf{t}_{rs}^{C_{t}} \
\end{matrix}
\right]^{-1}
\mathbf{X}_{gs}^{C_{t}}  \right)} = 0.
\end{equation}
Finally, we obtain the RS essential matrix whose DOF is 17: one GS transformation (rotation/translation) and two RS transformations (angular/linear velocities) up to scale $(3+3-1+(3+3)\times2=17)$ as
\begin{equation}
\mathbf{E}_{rs} 
= \left(\mathbf{R}_{rs}^{C_{t-1}}\right)^{T}  \mathbf{R}_{gs}\left\lfloor \mathbf{t}_{gs} - \mathbf{t}_{rs}^{C_{t}}  + \mathbf{R}_{gs} \mathbf{t}_{rs}^{C_{t-1}}  \right\rfloor_{\times}    \mathbf{R}_{rs}^{C_{t}}  .
\end{equation}

\begin{algorithm}[tb]
	\caption{Monocular Rolling Shutter Visual Odometry (MRSVO)}
	\small
	
	\label{alg:MRSVO} 
	\SetKwInOut{Input}{Input}
	\SetKwInOut{Output}{Output}
	\Input{Feature Point Correspondences ($\mathbf{m}_{rs,t-1:t}^{i}$)}
	
	%	$\bar{\mathcal{B}} \leftarrow \emptyset$\;
	$[\hat{\mathbf{q}}_\text{init} , \hat{\mathbf{t}}_\text{init} ] \leftarrow \text{conventional transformation estimation}(\mathbf{m}_{rs,1:N}^{C_{t-1}},\mathbf{m}_{rs,1:N}^{C_{t}})$\;

	$k \leftarrow 1$, $n_\text{max} \leftarrow 0$, $itr \leftarrow 500$\;
	\While{$k \le \text{itr}$}
	{
		$\tilde{\mathbf{x}}_{init} \leftarrow \mathbf{0}_{18 \times 1}$\;
		
		$[ \mathbf{m}_{rs,i}^{C_{t-1}}, \mathbf{m}_{rs,i}^{C_{t-1}}]_{20pt} \leftarrow \text{random sampling}(\mathbf{m}_{rs,1:N}^{C_{t-1}},\mathbf{m}_{rs,1:N}^{C_{t}})$\;
		
		$(\tilde{\mathbf{x}}_{opt})_{18 \times 1} \leftarrow \text{LM algorithm} (\tilde{\mathbf{x}}_{init};\hat{\mathbf{q}}_\text{init} , \hat{\mathbf{t}}_\text{init}, [ \mathbf{m}_{rs,i}^{C_{t-1}}, \mathbf{m}_{rs,i}^{C_{t-1}}]_{20pt} )
		$

		$\left(\hat{\mathbf{x}}_{opt}\right)_{19 \times 1} \leftarrow
		\text{nominal state conversion} (\hat{\mathbf{q}}_{\text{init}}, \hat{\mathbf{t}}_\text{init},\tilde{\mathbf{x}}_{opt})$
		
		$n_\text{inlier} \leftarrow$ $\text{count inliers}$($\hat{\mathbf{x}}_{opt}$)\;
		
		\If{ $n_\text{inlier}$ $\ge n_\text{max}$}
		{
			$[\hat{\mathbf{q}}_\text{gs}, \hat{\mathbf{t}}_\text{gs}, \hat{\mathbf{w}}^{C_{t-1}}, \hat{\mathbf{v}}^{C_{t}}, \hat{\mathbf{w}}^{C_{t-1}}, \hat{\mathbf{v}}^{C_{t}}]_{final} \leftarrow \hat{\mathbf{x}}_{opt}$\;
			$n_\text{max} \leftarrow  n_\text{inlier}$
		}            
		$k \leftarrow k+1$\;   
	}

	\Output{GS Transformation ($ \hat{\mathbf{q}}_\text{gs}$, $\hat{\mathbf{t}}_\text{gs}$) \\ \ and Instantaneous RS Motion (~$\hat{\mathbf{w}}^{C_{t-1}}$,~$\hat{\mathbf{v}}^{C_{t}}$,~$\hat{\mathbf{w}}^{C_{t-1}}$, ~$\hat{\mathbf{v}}^{C_{t}}$)  }

\end{algorithm}

\vspace{-2mm}
\section{ Estimation of Rolling Shutter Camera Motion  } \label{sec:motion_estimation}

% nonlinear problem
The conventional essential matrix estimation is performed by Direct Linear Transformation (DLT), and rotation $\mathbf{R}_{gs}$ and translation $\mathbf{t}_{gs}$ are extracted by Singular Value Decomposition (SVD) \cite{Hartley:BOOK:2003}.
As described in Sec.~\ref{sec:intro}, the conventional motion estimation becomes inaccurate with RS images.
Thus, we jointly estimate the GS and RS camera motion.
The overall process of the proposed Monocular Rolling Shutter Visual Odometry (MRSVO) is described in Algorithm~\ref{alg:MRSVO}. 
The constraint Eq.~(\ref{eq:rs_essential_matrix}) between RS feature points is highly nonlinear.
For that reason, we estimate the solution of this nonlinear equation with the Levenberg-Marquardt algorithm.

From the RS essential matrix, the nominal state variables to estimate $\mathbf{x} \in \mathbb{R}^{19}$ are defined as
\begin{equation}
\mathbf{x} = 
\left[ \ \mathbf{q}_{gs}, \ \mathbf{t}_{gs}, \ \mathbf{w}^{C_{t-1}}, \ \mathbf{v}^{C_{t-1}}, \ \mathbf{w}^{C_{t}}, \ \mathbf{v}^{C_{t}} \ \right]^{T},
\end{equation}
where $\mathbf{q}_{gs} \in \mathbb{R}^{4}$ denotes a quaternion which expresses the 3-DOF rotation $\mathbf{R}_{gs}$.
%
%The proposed RS essential matrix consists of the composition of three coordinate transformation: 1 for two GS images and  2 for RS-GS images in each frame.

The error states $\tilde{\mathbf{x}} \in \mathbb{R}^{18 \times 1}$  inspired from \cite{Maybeck:Book:1982} is introduced to reduce the complexity as
\begin{equation}
\tilde{\mathbf{x}} = 
\left[ \ \mathbf{\delta \theta}_{gs} \ \delta \mathbf{t}_{gs} \ \delta \mathbf{w^{C_{t-1}}} \ \delta \mathbf{v^{C_{t-1}}} \ \delta \mathbf{w^{C_{t}}} \ \delta \mathbf{v^{C_{t}}} \ \right],
\end{equation}
where $\mathbf{\delta\theta}_{gs} \in \mathbb{R}^{3 \times 1}$ is the angular error of GS rotation which is originated from error quaternion $\delta \mathbf{q} \simeq \left[ \  1 \ \  \frac{1}{2} \delta \theta^{T}  \ \right]^{T}$.

The initial nominal states $\mathbf{q}_{gs}, \ \mathbf{t}_{gs}$ are obtained from the conventional DLT and SVD algorithm. 
%
%The initial nominal RS states $ \mathbf{w^{C_{t-1}}}, \  \mathbf{v^{C_{t-1}}}, \  \mathbf{w^{C_{t}}}, \  \mathbf{v^{C_{t}}} $ can be obtained by interpolating nearest frames.
%
We set the initial state to zero to handle arbitrary RS distortions. 

The cost function is constructed as the Sampson error as
\begin{equation}
\small
\begin{aligned}
\min_{\tilde{\mathbf{x}}} { \sum_{ i }^{ 17 }{ \frac{ \left\| (\mathbf{m}_{rs,i}^{C_{t}})^{T} \mathbf{F}_{rs}( \tilde{\mathbf{x}} )  \mathbf{m}_{rs,i}^{C_{t-1}} \right\|_{2}^{2} }{\left( \mathbf{F}_{rs}(\tilde{\mathbf{x}})\mathbf{m}^{C_{t-1}}_{rs,i}\right)_{c}^{2} + \left( \mathbf{F}_{rs}(\tilde{\mathbf{x}})\mathbf{m}^{C_{t-1}}_{rs,i}\right)_{r}^{2} + \left( \mathbf{F}_{rs}(\tilde{\mathbf{x}})\mathbf{m}^{C_{t}}_{rs,i}\right)_{c}^{2} + \left( \mathbf{F}_{rs}(\tilde{\mathbf{x}})\mathbf{m}^{C_{t}}_{rs,i}\right)_{r}^{2} } } }, \\
s.t \ \ \mathbf{F}_{rs}\left( \tilde{\mathbf{x}}\right) = \mathbf{K}^{-T} \mathbf{E}_{rs} \left( \tilde{\mathbf{x}}\right) \mathbf{K}^{-1}.
\end{aligned}
\end{equation}
where the subscript $c$ and $r$ indicate the column and the row of a feature point in the normalized camera coordinate, respectively.

% ransac
The nonlinear least square methods such as the LM algorithm is easy to converge to local minimum.
To avoid bad initialization as well as outliers, we perform the RANSAC process. % with 15 points according to the DOF of the RS fundamental matrix.
We give 20 feature point correspondences as the input of the RANSAC for robustness.

\begin{figure}[tb]
	\centering
	\subfloat[3D view]{\includegraphics[width=0.53\linewidth]{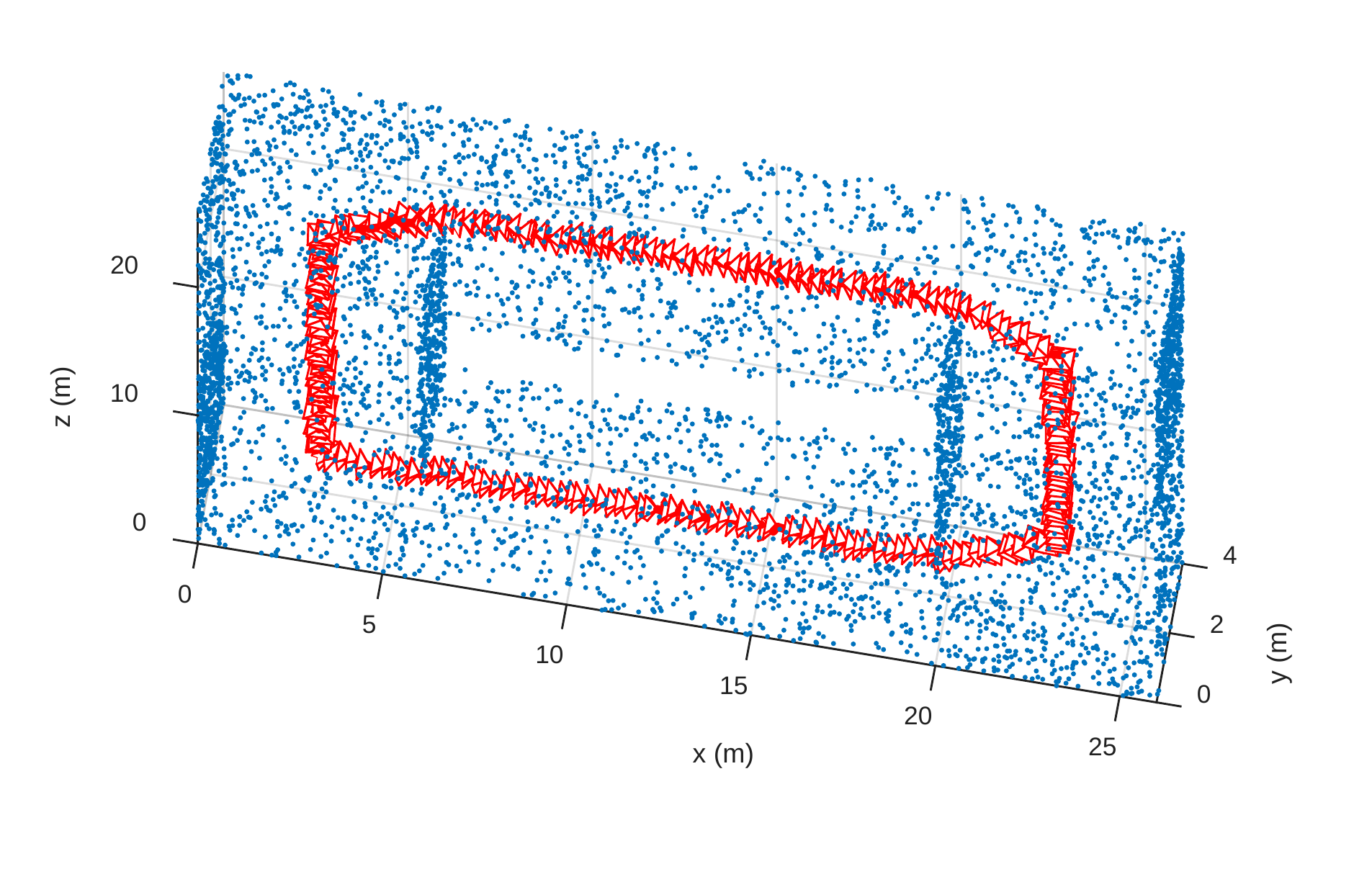}}
	\subfloat[Top view]{\includegraphics[width=0.40\linewidth]{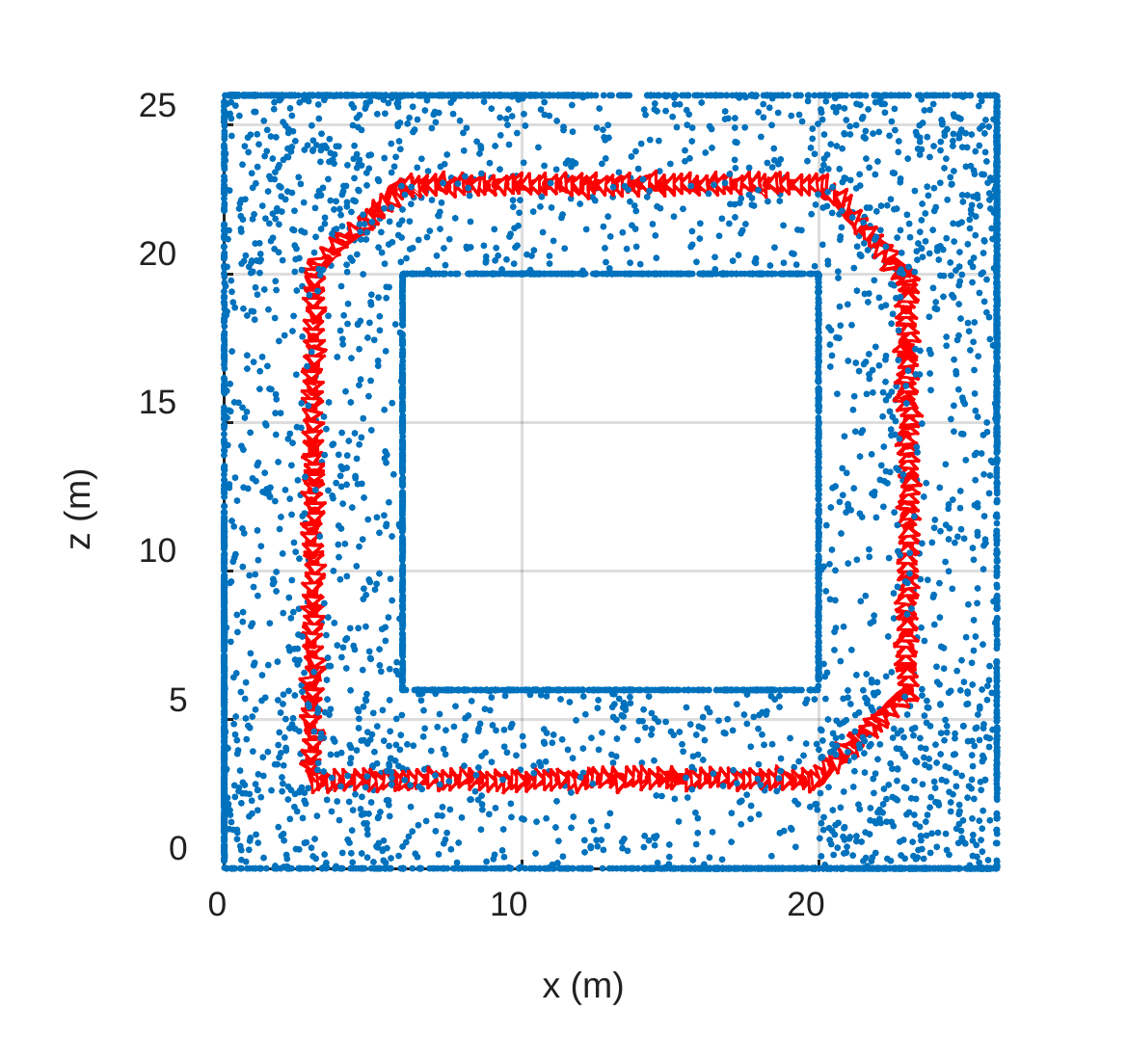}}	
	\caption{Synthesized data for ego-motion estimation. Blue dots represent 3D landmarks, and red polyhedrons represent the trajectory of moving camera. The camera mainly moves along x- and z-axes, and start- and end-points are the same as $\left[ 2, 2, 2 \right]^{T}(m)$.}		
	\label{fig:synthetic_data}
	\vspace{-4mm}
\end{figure}

\section{Experimental Results} \label{sec:experiments}

To validate the proposed RS camera motion estimation algorithm, we perform several experiments on synthetic dataset as well as real dataset. 
%
%Since there is no previous work on the online ego-motion estimation with a monocular RS camera to the best of our knowledge, 
Since there is no previous work that specifically focuses on the RS distortion for the online ego-motion estimation to the best of our knowledge, 
we compare the proposed algorithm with the conventional MVO based on the fundamental matrix with the normalized 8-point algorithm \cite{Hartley:BOOK:2003}. 
The quantitative and qualitative comparisons on both datasets explicitly show the superiority of proposed algorithm.
In the synthetic dataset, we focus on analyzing the influence of feature tracking error due to RS effects on relative motion estimates with two consecutive RS images.
It empirically reveals that the inlier ratio and the accuracy of relative motion estimates are highly correlated.
In the real dataset, we verify the superiority of our algorithm in several sequences taken by a hand-held RS camera. 

\begin{figure}[tbp]
	\centering	\vspace{-8pt}
	\subfloat[Noise-free ]{\label{fig:syn_noise_free_exp}\includegraphics[width=0.325\linewidth]{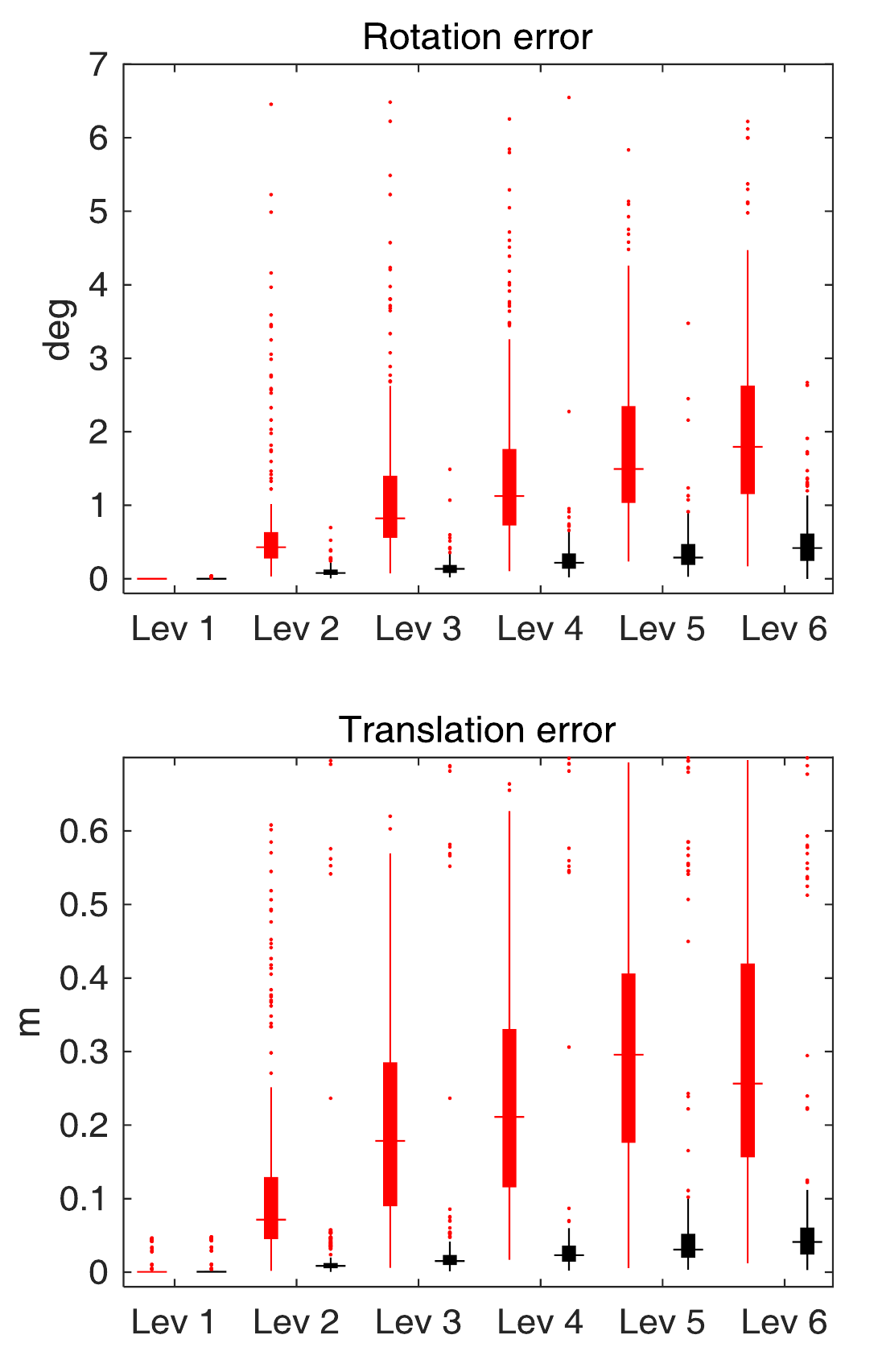}}
	\subfloat[Gaussian noise ]{\label{fig:syn_noise_gaussian_exp}\includegraphics[width=0.33\linewidth]{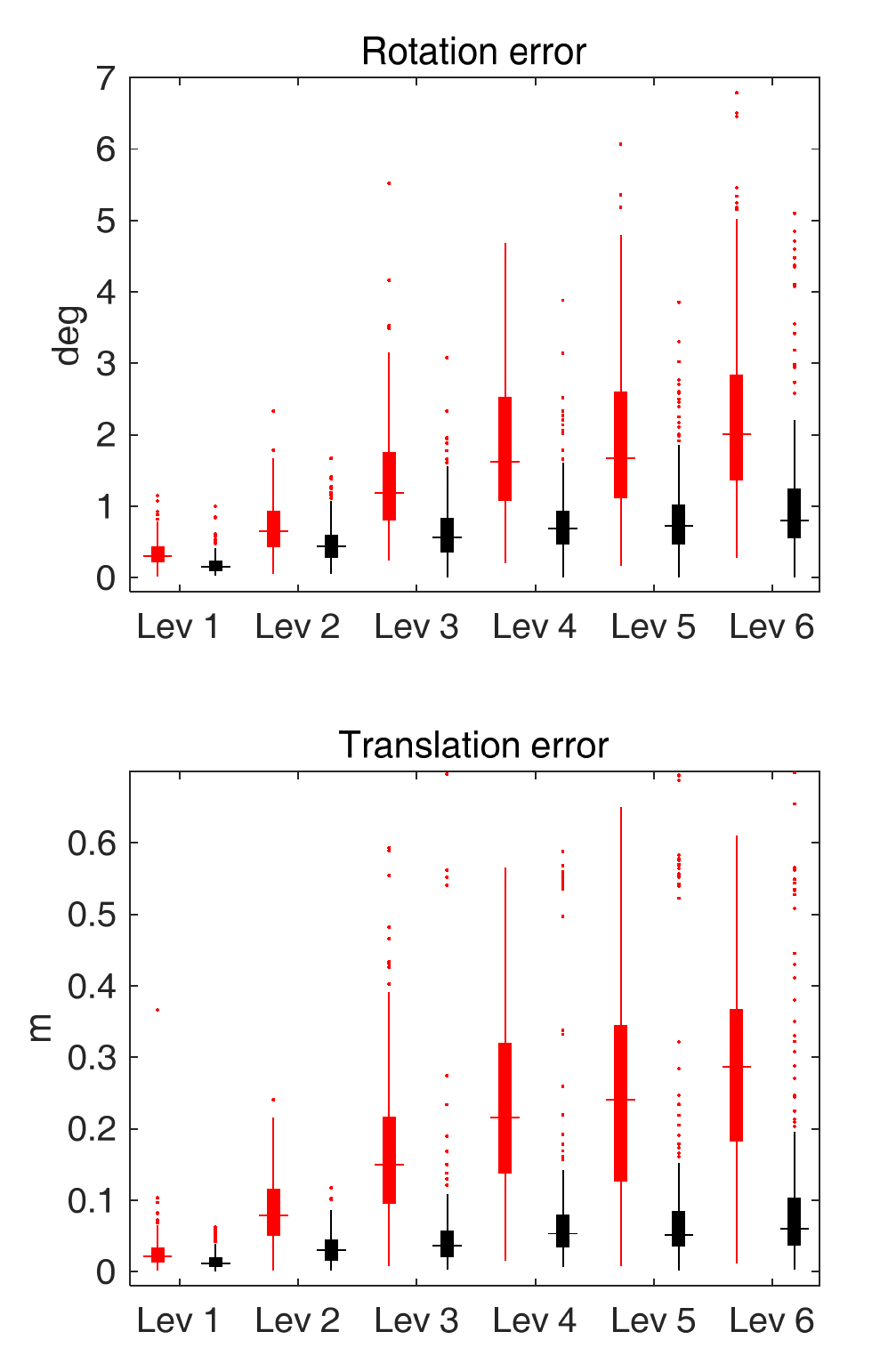}}		
	\subfloat[Laplacian noise ]{\label{fig:syn_noise_laplacian_exp}\includegraphics[width=0.32\linewidth]{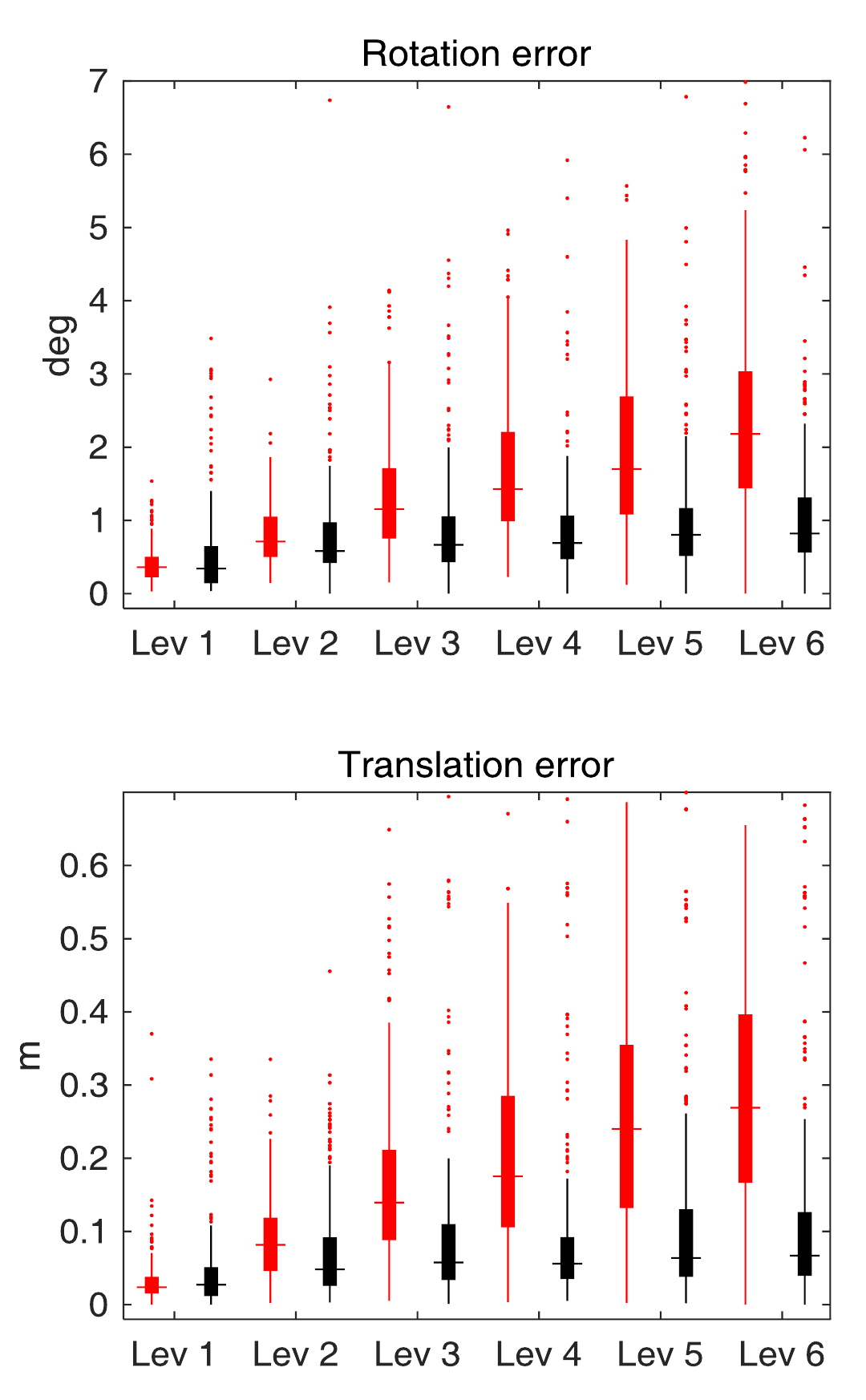}}			
	%	\subfigure[Gaussian noise ]{\label{fig:exp_with_gaussian_noise}\includegraphics[width=0.3\linewidth]{./img/exp_with_gaussian_noise_v2}}		
	%	\subfigure[Laplacian noise ]{\label{fig:exp_with_laplacian_noise}\includegraphics[width=0.3\linewidth]{./img/exp_with_laplacian_noise_v1}}		
	\caption{Comparison of the MVO and the proposed MRSVO on three types of noisy environments. The error statistics are illustrated with the box-and-whisker diagram. The lines in the middle of the bars indicate the median value of the errors. The red dots  are outliers. The results of the MVO are expressed by red color and the results of the MRSVO are expressed by black color. }		
	\vspace{-4mm}
\end{figure}

\vspace{0mm}
\subsection{Synthetic data}

% synthetic data generation
We generate synthetic data with a smooth trajectory and 3D points as shown in Fig. \ref{fig:synthetic_data} .
The trajectory is manually specified and interpolated with a spline function, and then we put the random noise to simulate realistic camera motion.
Traveling distance is about 80 $(m)$.
All visible 3D points are projected onto the image.
The 250 image are captured at 5 Hz.
The resolution of the image is 1280 $\times$ 720 and the focal length is 5 mm.
The number of feature points is limited up to 500 with a bucketing strategy to obtain evenly distributed feature points.
In addition, arbitrary RS distortion is added to 2D feature points in every image.
The RS distortion depends on one-line exposure time $\tau$ of an RS camera, instantaneous linear and angular velocity $\mathbf{v}_{rs},\mathbf{w}_{rs}$ in a frame.
In our experiments, we fix $\tau$ to 50 $\mu s$ and make $\mathbf{v}_{rs},\mathbf{w}_{rs}$ change from 0 to [50,100] $(m/s,deg/s)$ since $\tau$ and $\mathbf{v}_{rs},\mathbf{w}_{rs}$ are inverse-correlated.
The RS distortion is generated at 6 levels ([0,0], [10,20], [20,40], [30,60], [40,80], [50,100] $(m/s,deg/s)$), and the estimates are evaluated with 100 Monte Carlo simulations.
The error is evaluated every two frames, and they are averaged for the whole sequence.
Rotation and translation errors are evaluated with metrics defined as 
\begin{equation}
\delta \mathbf{T} 
= 
{\left[
	\begin{matrix}
	\hat{\delta\mathbf{R}} & {\delta\mathbf{t}} \\
	\mathbf{0}_{3 \times 1} & 1
	\end{matrix}
\right]} 
= 
\mathbf{T}_{gt} \mathbf{T}_{est}^{-1}, \ \
\delta \theta = f \left( \hat{\delta\mathbf{R}}\right) , \ \
\delta t = \left\| \delta\mathbf{t} \right\|_{2} ,
\end{equation}
where $f$ is Rodrigues' rotation formula.

The conventional MVO and the proposed MRSVO estimate relative motion up to scale.
Therefore, we utilize the ground-truth scale information for accurate evaluation. 
Furthermore, for the convenience of the comparison, we name the conventional MVO which estimates ego-motion with the essential matrix \cite{Hartley:BOOK:2003} simply as MVO.

{\renewcommand{\arraystretch}{0.92}
	\begin{table}[tb]
		\centering
		\scriptsize
		\renewcommand{\tabcolsep}{1.5mm}		
		\subfloat[Noise-free]{
			\centering
			\label{tab:exp_without_noise}
			\begin{tabular}{c|c|c|cccccc}
				\hline
				\multicolumn{3}{c}{}                                                       & Lev 1       & Lev 2       & Lev 3       & Lev 4       & Lev 5       & Lev 6       \\ \hline
				\multirow{4}{*}{Mean}               & \multirow{2}{*}{Rotation(deg)}    & MVO & 0.000   & 0.686   & 1.266   & 1.478   & 1.805   & 2.093   \\ 
				&                              & MRSVO & 0.007   & 0.036   & 0.041   & 0.052   & 0.373   & 0.475   \\
				& \multirow{2}{*}{Translation (m)} & MVO & 0.007   & 0.119   & 0.203   & 0.233   & 0.296   & 0.284   \\
				&                              & MRSVO & 0.012   & 0.026   & 0.038   & 0.053   & 0.085   & 0.078   \\ \hline
				\multirow{4}{*}{\begin{tabular}[c]{@{}c@{}}Standard\\   deviation\end{tabular}} & \multirow{2}{*}{Rotation(deg)}    & MVO  & 0.000   & 0.889   & 1.389   & 1.116   & 1.074   & 1.513   \\
				&                              & MRSVO & 0.009   & 0.076   & 0.137   & 0.448   & 0.341   & 0.368   \\
				& \multirow{2}{*}{Translation(m)} & MVO & 0.015   & 0.129   & 0.141   & 0.147   & 0.157   & 0.160   \\
				&                              & MRSVO & 0.015   & 0.121   & 0.116   & 0.125   & 0.167   & 0.139   \\ \hline				
				\multicolumn{2}{c|}{\multirow{2}{*}{Inlier ratio (\%)}}     & MVO & 100.0  & 44.9  & 34.4  & 30.2  & 27.4  & 25.4  \\ 
				\multicolumn{2}{c|}{}                                  & MRSVO & 100.0  & 99.9  & 99.9  & 99.5  & 97.9  & 97.2  \\ \hline	
				%				\multicolumn{2}{c|}{\multirow{2}{*}{The number of inliers}}         & MVO & 500.000 & 224.594 & 172.028 & 150.980 & 137.020 & 127.141 \\
				%				\multicolumn{2}{c|}{}                                               & MRSVO & 500.000 & 499.731 & 499.349 & 497.382 & 489.474 & 486.080 \\ \hline
			\end{tabular}
		}\\
		\subfloat[Gaussian noise]{
			\centering
			\label{tab:exp_with_noise_gaussian}
			\begin{tabular}{c|c|c|cccccc}
				\hline
				\multicolumn{3}{c}{}                                                       & Lev 1       & Lev 2       & Lev 3       & Lev 4       & Lev 5       & Lev 6       \\ \hline 
				\multirow{4}{*}{Mean}                                                           & \multirow{2}{*}{Rotation (deg)}  & MVO & 0.349   & 0.704   & 1.371   & 1.844   & 1.955   & 2.252   \\
				&                                  & MRSVO      & 0.186   & 0.482   & 0.657   & 0.766   & 0.859   & 1.045   \\
				& \multirow{2}{*}{Translation (m)} & MVO & 0.027   & 0.085   & 0.167   & 0.235   & 0.249   & 0.282   \\
				&                                  & MRSVO      & 0.017   & 0.033   & 0.055   & 0.097   & 0.100   & 0.106   \\ \hline
				\multirow{4}{*}{\begin{tabular}[c]{@{}c@{}}Standard\\   deviation\end{tabular}} & \multirow{2}{*}{Rotation (deg)}  & MVO & 0.189   & 0.346   & 0.778   & 0.994   & 1.111   & 1.244   \\
				&                                  & MRSVO      & 0.134   & 0.278   & 0.413   & 0.499   & 0.608   & 0.890   \\
				& \multirow{2}{*}{Translation (m)} & MVO & 0.028   & 0.047   & 0.103   & 0.131   & 0.148   & 0.138   \\
				&                                  & MRSVO      & 0.014   & 0.021   & 0.088   & 0.140   & 0.148   & 0.131   \\ \hline
				\multicolumn{2}{c|}{\multirow{2}{*}{Inlier ratio(\%)}}     & MVO & 23.2  & 21.5  & 19.4  & 18.4  & 18.0  & 17.4  \\
				\multicolumn{2}{c|}{}                                  & MRSVO & 51.9  & 50.0  & 49.2  & 48.5  & 47.9  & 47.7  \\ \hline				
				%				\multicolumn{2}{c|}{\multirow{2}{*}{The number of inliers}}                                                         & MVO & 116.137 & 107.422 & 97.068  & 91.803  & 90.197  & 87.048  \\
				%				\multicolumn{2}{c|}{}                                                                                               & MRSVO      & 259.486 & 250.124 & 245.884 & 242.606 & 239.369 & 238.353 \\ \hline
			\end{tabular}
		}\\
		\subfloat[Laplacian noise]{
			\centering
			\label{tab:exp_with_noise_laplacian}
			\begin{tabular}{c|c|c|cccccc}
				\hline
				\multicolumn{3}{c}{}                                                       & Lev 1       & Lev 2       & Lev 3       & Lev 4       & Lev 5       & Lev 6       \\ \hline
				\multirow{4}{*}{Mean}                                                           & \multirow{2}{*}{Rotation (deg)}  & MVO & 0.401   & 0.805   & 1.337   & 1.675   & 2.017   & 2.421    \\
				&                                  & MRSVO      & 	0.541   & 0.820   & 0.922   & 0.930   & 1.064   & 1.066   \\
				& \multirow{2}{*}{Translation (m)} & MVO & 0.032   & 0.089   & 0.168   & 0.204   & 0.257   & 0.284   \\
				&                                  & MRSVO      & 0.047   & 0.073   & 0.100   & 0.096   & 0.116   & 0.121   \\ \hline
				\multirow{4}{*}{\begin{tabular}[c]{@{}c@{}}Standard\\   deviation\end{tabular}} & \multirow{2}{*}{Rotation (deg)}  & MVO & 0.246   & 0.427   & 0.813   & 0.943   & 1.270   & 1.370   \\
				&                                  & MRSVO      & 0.641   & 0.741   & 0.865   & 0.875   & 1.015   & 0.853   \\
				& \multirow{2}{*}{Translation (m)} & MVO & 0.036   & 0.057   & 0.118   & 0.132   & 0.150   & 0.147   \\
				&                                  & MRSVO      & 0.061   & 0.071   & 0.122   & 0.123   & 0.138   & 0.148   \\ \hline
				\multicolumn{2}{c|}{\multirow{2}{*}{Inlier ratio(\%)}}     & MVO & 9.4   & 8.7   & 8.0   & 7.6   & 7.4   & 7.0   \\
				\multicolumn{2}{c|}{}                                  & MRSVO & 55.2  & 54.7  & 53.4  & 53.2  & 52.9  & 50.9  \\ \hline
				%				\multicolumn{2}{c|}{\multirow{2}{*}{The number of inliers}}                                                         & MVO & 47.133  & 43.622  & 40.096  & 37.892  & 36.755  & 35.189  \\
				%				\multicolumn{2}{c|}{}                                                                                               & MRSVO      & 276.193 & 273.727 & 266.863 & 266.116 & 264.502 & 254.526 \\ \hline
			\end{tabular}
		}	
		\caption{Error statistics of the MVO and the MRSVO on three types of noise environments.}
	\vspace{-5mm}
	\end{table}

}

% noise free experiemtns
\subsubsection{Comparison without feature tracking noise}

To validate the effect of the proposed algorithm on RS distortion, we evaluate the MRSVO with true 2D feature correspondences.
Figure \ref{fig:syn_noise_free_exp} shows that the MRSVO outperforms the MVO. 
Rotation and translation errors of both the MVO and the MRSVO linearly increase as the level of RS distortion increases.
However, the rate of change of the MRSVO is much smaller than that of the MVO.
More detailed results are described in Table \ref{tab:exp_without_noise} with the average inlier ratio.
At level 1 (no RS distortion), the inlier ratios of the MRSVO and the MVO are 100$\%$ when the number of tracked points is 500.
As the level of the RS distortion increases, the inlier ratio of the MVO dramatically decreases  (up to 25.4$\%$).
On the contrary, the inlier ratio of the MRSVO is reduced little (above 97.2$\%$), and 
the proposed MRSVO with high inlier ratios shows the lower estimation error. % in identical noise environments.
Consequently, we can evaluate the performance of the MRSVO using the inlier ratio in real dataset.

\begin{figure}[tb]
	\centering
	\includegraphics[width=0.98\linewidth]{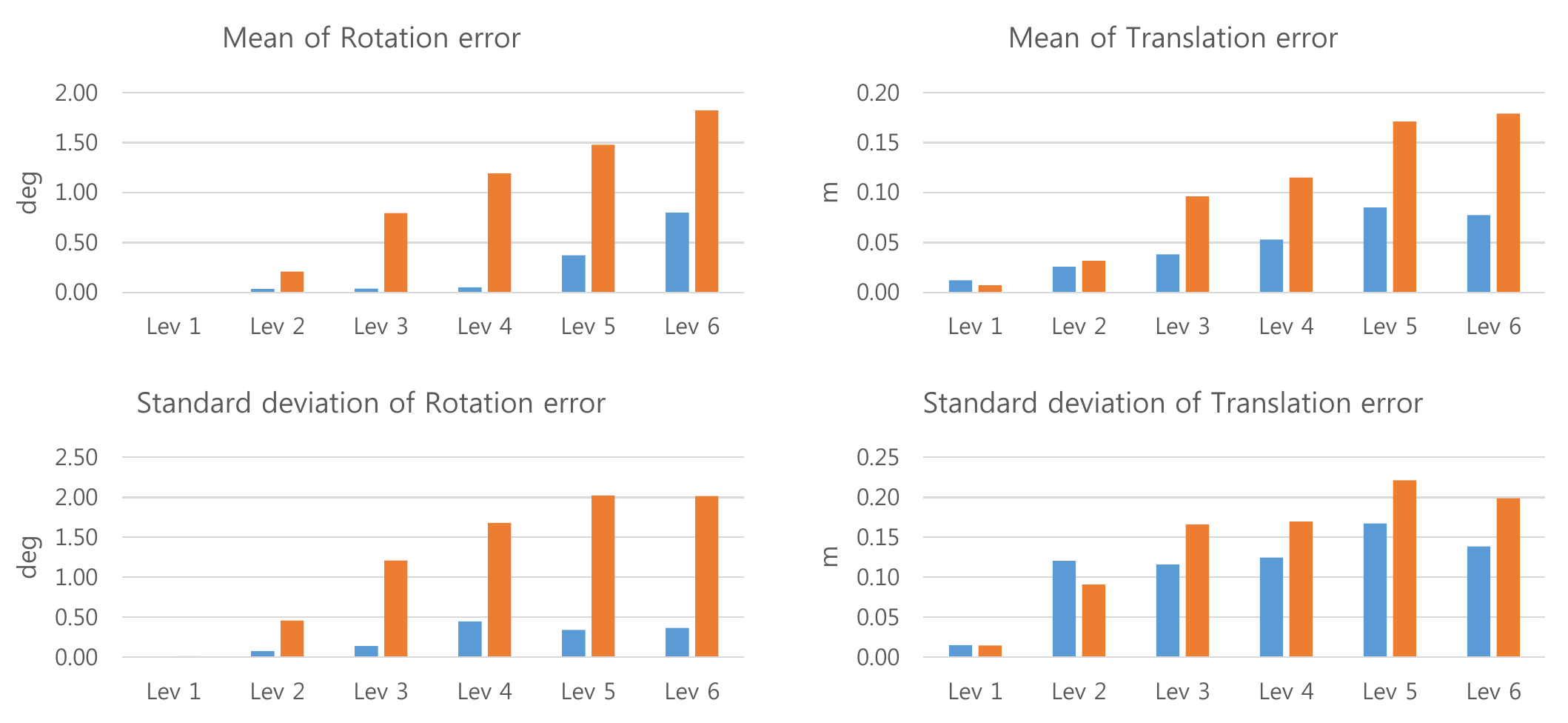}	
	\caption{Self comparison on the noise-free synthetic dataset. The results of the MRSVO with RANSAC are represented in the blue color, and the results of the MRSVO without RANSAC are represented in the orange color.}		
	\label{fig:syn_self_comparison_ransac}
	\vspace{-2mm}
\end{figure}

Even though the input feature points are noise-free, the high level of RS distortion degrades the MRSVO because the MRSVO exploits the estimate of the MVO as initial states.
To handle this, we apply the RANSAC process to avoid bad initialization as described in Sec.~\ref{sec:motion_estimation}.
Figure \ref{fig:syn_self_comparison_ransac} shows the performance of the MRSVO with RANSAC and without RANSAC.
The effect of RANSAC is clearly shown in higher levels of RS distortion as expected.

\subsubsection{Comparison with feature tracking noise}

We evaluate the MRSVO and the MVO under Gaussian and Laplacian noises.
The standard deviation of two types of noise is set to 1.
The randomly generated noises are added to positions of 2D feature point correspondences.
Figure~\ref{fig:syn_noise_gaussian_exp} and Fig.~\ref{fig:syn_noise_laplacian_exp} show that the MRSVO outperforms the MVO under both types of noise.
The estimation errors show a similar tendency with the noise-free case.
Table \ref{tab:exp_with_noise_gaussian} and \ref{tab:exp_with_noise_laplacian} describe the mean and standard deviation values of errors and the average inlier ratio.
The inlier ratios of both the MRSVO and the MVO in the Gaussian and Laplacian noise environments are decreased compared to the noise-free case.
In the Gaussian noise case, both the mean and the standard deviation of the MRSVO are lower than those of the MVO at all the levels of the RS  distortion. 
Besides, we can notice that the MRSVO produces more accurate estimates and higher inlier ratios than the MVO at level 1 (no RS distortion) in the Gaussian noise case.
It means that the MRSVO effectively suppresses the noise of feature points as well as RS distortion.
In the case of Laplacian noise, the MRSVO outperforms the MVO in the high levels (4-6) of the RS distortion. 
However, the MVO provides slightly more accurate estimates than the MRSVO in the low levels (1-3)  of the RS distortion, because a large number of outliers owing to the Laplacian noise sometimes lead to wrong convergence of the LM algorithm.

\begin{figure}[tb]
	\centering	
	\includegraphics[width=0.99\linewidth]{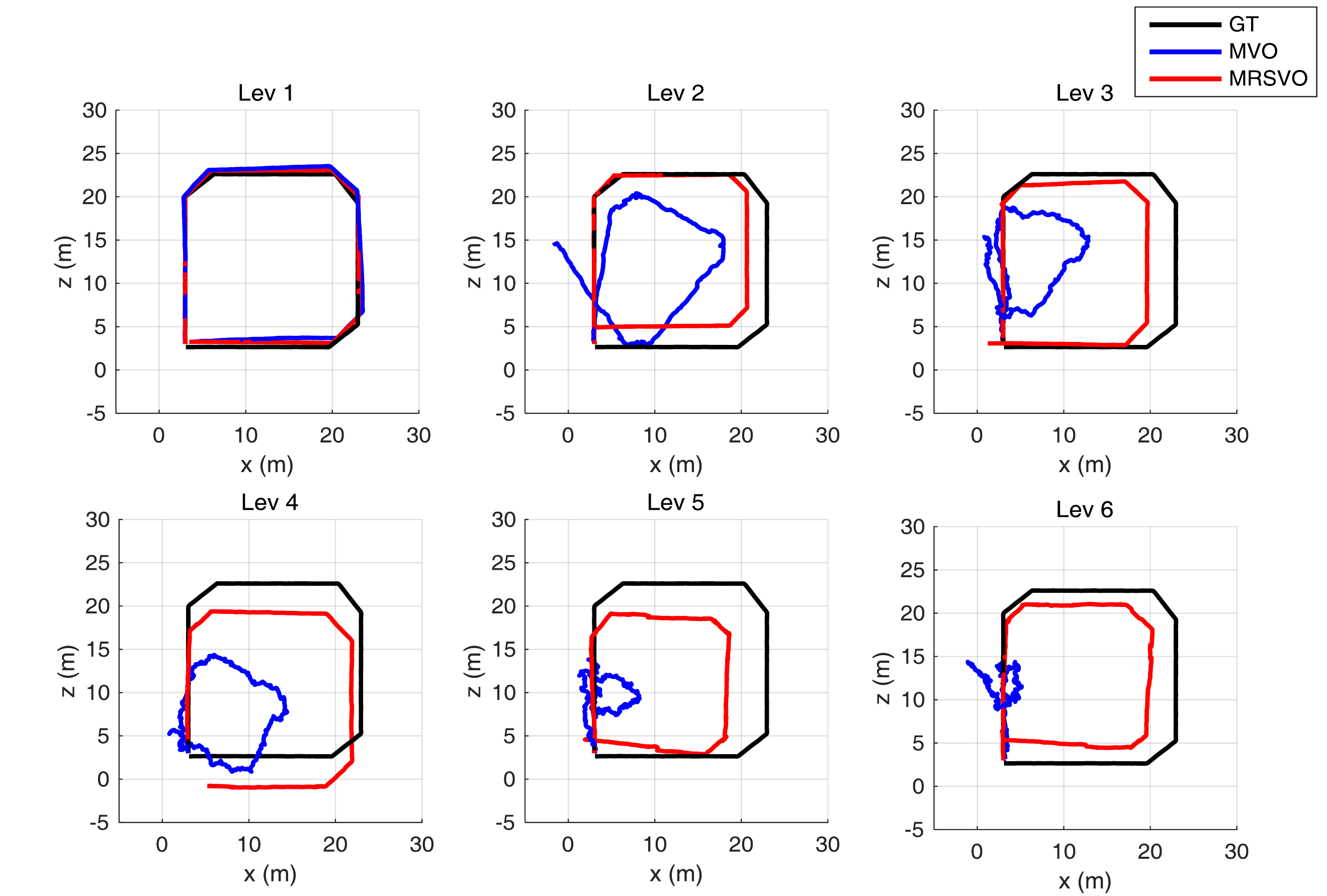}
	\caption{Trajectories constructed by the MVO and the MRSVO with different levels of RS distortions. }		
	\label{fig:trajectory}
	\vspace{-4mm}
\end{figure}

The relative motion estimates from two consecutive frames are concatenated to construct the motion trajectory as 
\begin{equation}
\mathbf{T}_{k+1} = \mathbf{T}_{k} 
{\left[
	\begin{matrix}
	\hat{\mathbf{R}}_{gs} & \hat{\mathbf{t}}_{gs} \\
	\mathbf{0}_{3 \times 1} & 1
	\end{matrix}
	\right]}^{-1}
, \  \ 
\mathbf{T}_{0} = \mathbf{I}_{4 \times 4}.
\end{equation}
Figure~\ref{fig:trajectory} shows camera trajectories with six levels of RS distortion with Laplacian noise of feature points.
In the high levels of distortion, the MRSVO produces more stable trajectories compare to the MVO.

\subsection{Real dataset}

\begin{figure}[tb]
	\centering	
	\subfloat[ Sequence 2 (long) ]{\includegraphics[width=0.58\linewidth]{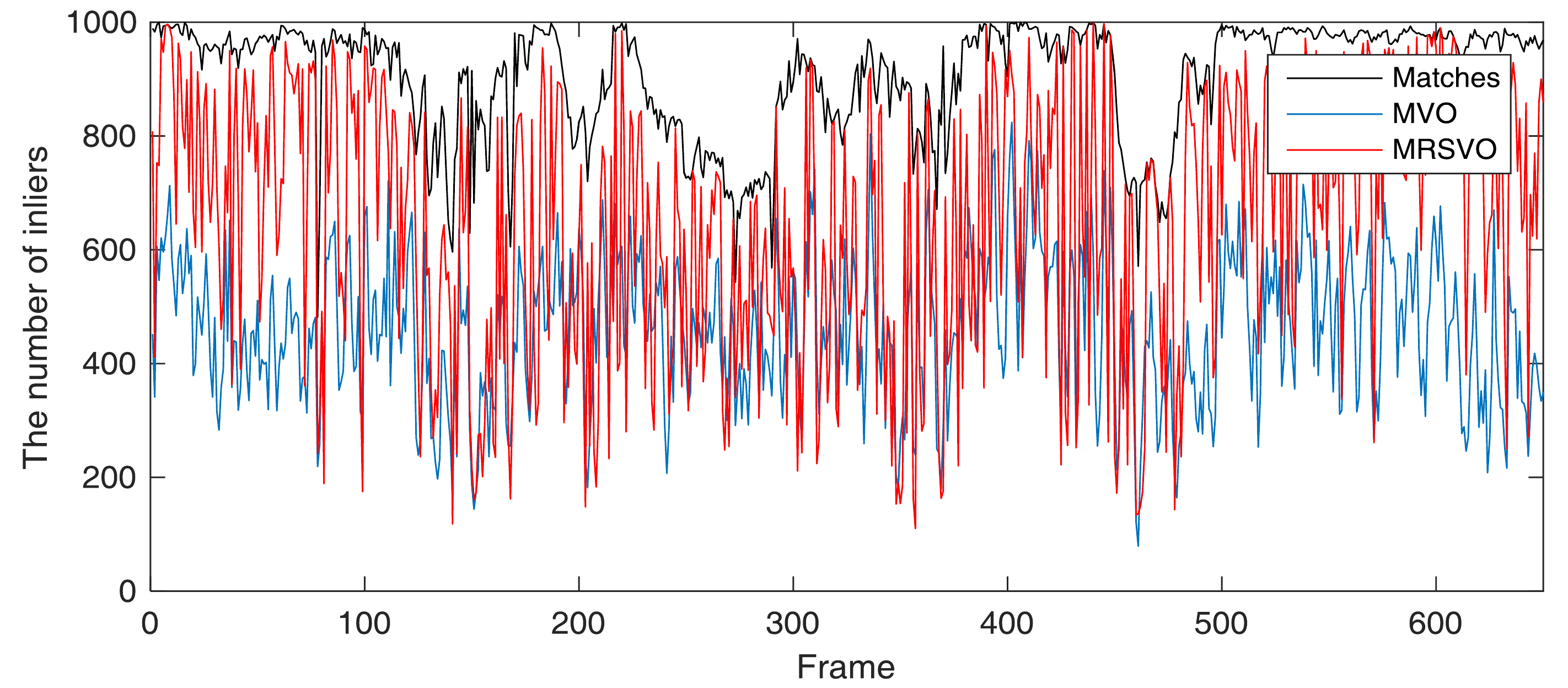}}
	\subfloat[ Sequence 11 (short)]{\includegraphics[width=0.40\linewidth]{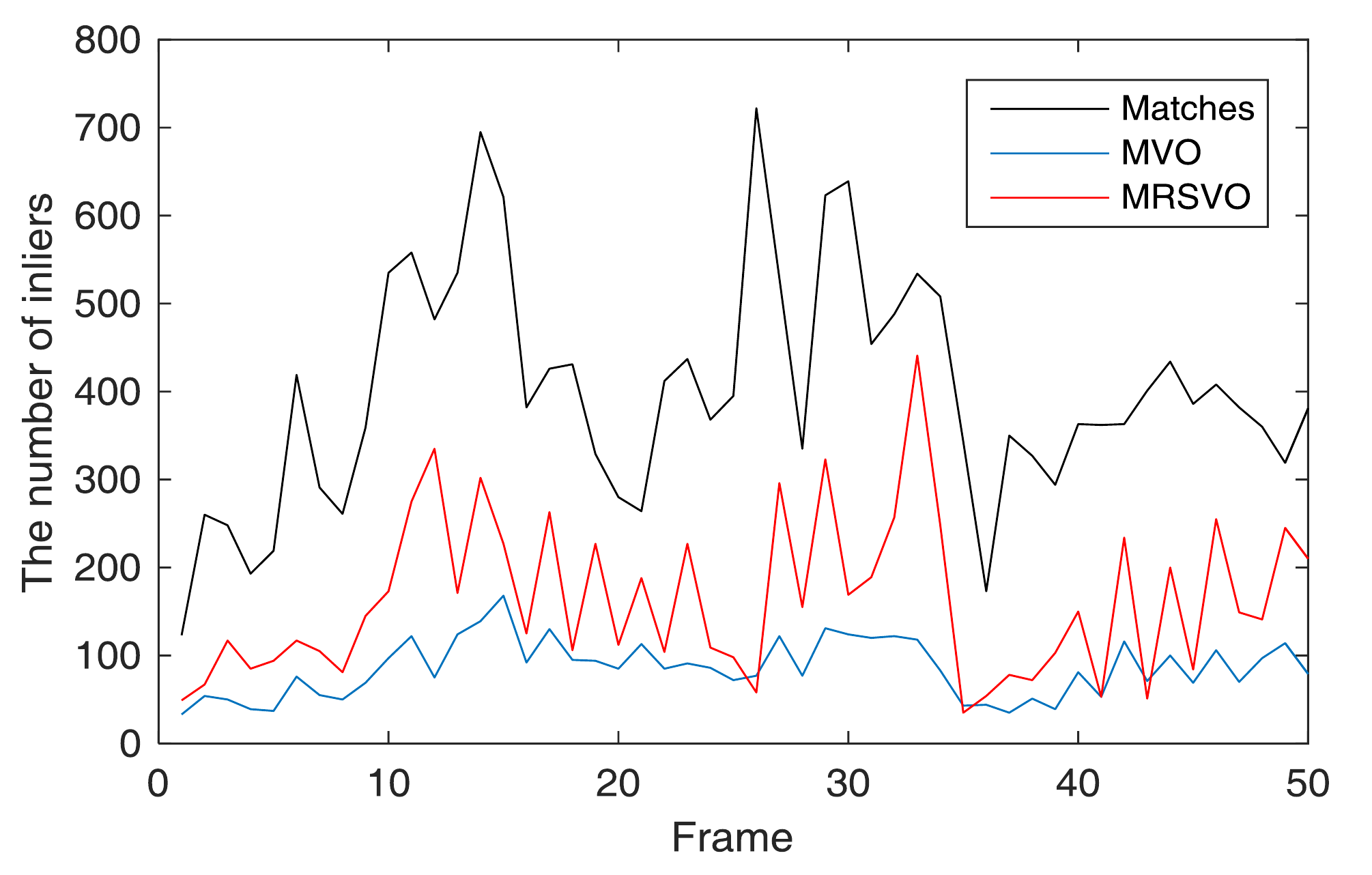}}	
	\caption{Number of inliers in selected sequences among 10 test sequences. }		
	\label{fig:real_history_inlier}
\end{figure}

{ 
	\renewcommand{\arraystretch}{1.1}
	\begin{table}[tb]
		\centering
		\scriptsize
		\renewcommand{\tabcolsep}{1.0mm}
		\caption{Average inlier ratios of the MVO and the MRSVO  on real RS dataset. $\text{MVO}^{*}$ indicates the results of the MVO on real GS dataset.}
		\label{tab:real_quantitative_inlier_comparison}
		\begin{tabular}{c|cc|cccccccc|c}
			\hline
			Sequence    & 1      & 2      & 3      & 7      & 9      & 10     & 11     & 12     & 20     & 21     & All    \\ \hline
			Frame       & 445    & 655    & 68     & 44     & 45     & 130    & 50     & 82     & 75     & 60    & 1714   \\ \hline
			MVO  & 61.7\% & 50.5\% & 51.1\% & 54.4\% & 35.2\% & 54.4\% & 17.1\% & 48.7\% & 52.5\% & 46.9\% & 47.2\% \\ \hline
			MRSVO       & 75.5\% & 71.6\% & 67.9\% & 66.5\% & 49.1\% & 56.7\% & 32.6\% & 67.7\% & 67.4\% & 72.4\% & 62.9\% \\ \hline \hline
			MVO* & 69.6\% & 58.0\% & 56.6\% & 52.5\% & 44.4\% & 52.0\% & 31.2\% & 47.7\% & 57.0\% & 57.0\% & 52.5\% \\ \hline
		\end{tabular}
	\end{table}
}

\begin{figure}[h]
	\centering	
	\includegraphics[width=0.99\linewidth]{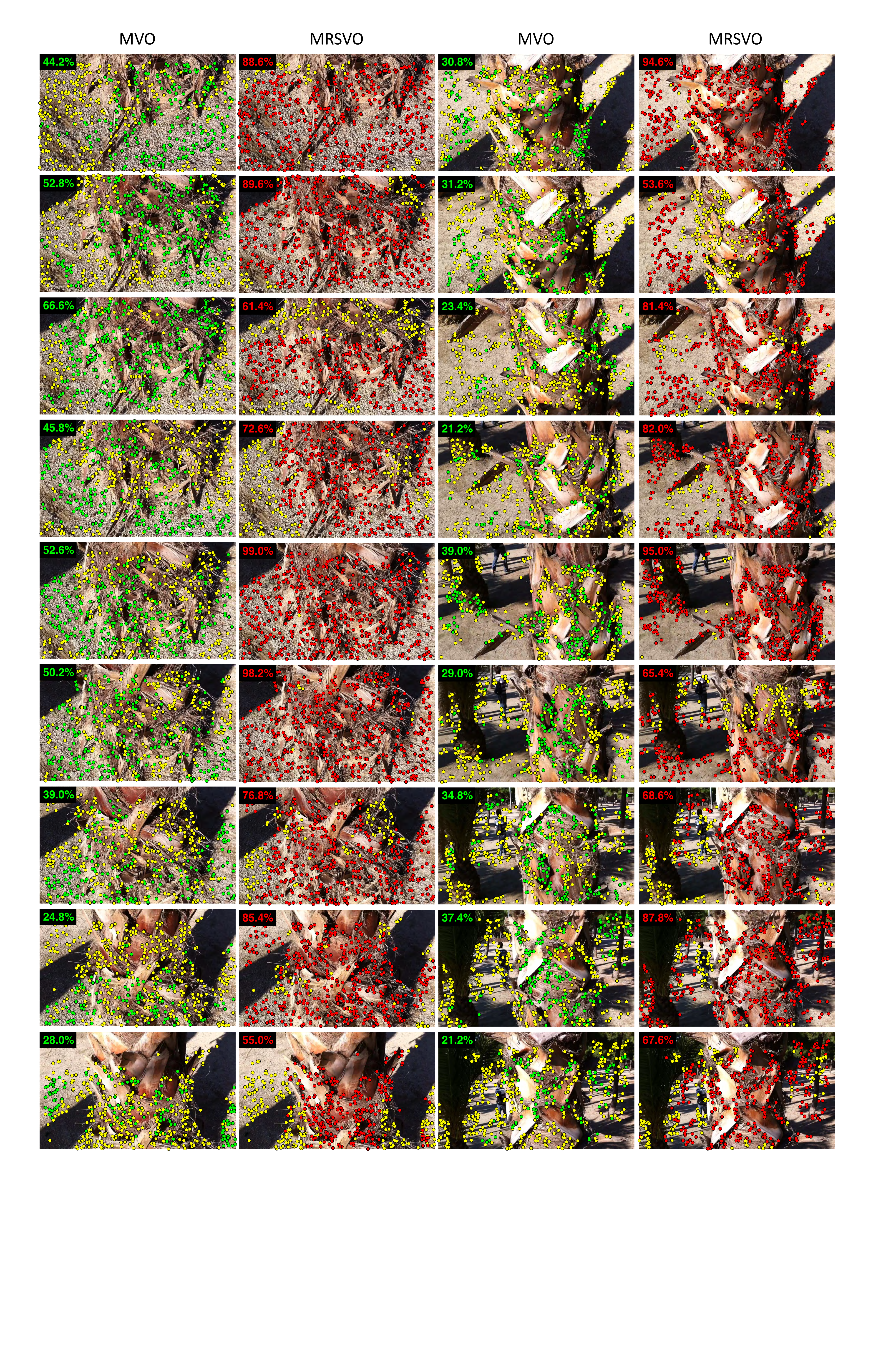}
	\caption{Qualitative comparison of the numbers of inliers of the MVO and the MRSVO on sequence 21. Green dots represent the inliers of the MVO, and red dots represent the inliers of the MRSVO. Yellow dots are outliers. We record the inlier ratio on the upper left corner of the each image. }		
	\label{fig:real_qualitative_inlier_comparison}
\end{figure}

We evaluate the MRSVO with the real dataset captured by commercial smart-phones cameras. 
%
%The experiments are performed in walking because the RS distortion appears in the hand-held cameras.
%
Here, we adopt  the inlier ratio as an evaluation metric since the inlier ratio is highly correlated with the estimation accuracy.

We compare the MRSVO to the MVO on the Hedborg's dataset \cite{Hedborg:CVPR:2012}.
The dataset contains 36 sequences which are composed of 2 long and 34 short videos.
The sequences were captured with iPhone 4 equipped with an RS camera and a cannon GS camera.   
We evaluate the MRSVO with selected 10 distinct sequences among 36 redundant sequences.
Table \ref{tab:real_quantitative_inlier_comparison} describes the average inlier ratios of the MRSVO and the MVO.
The MRSVO produces about 15$\%$ higher inlier ratio on average.
In the GS camera, the inlier ratio is increased about 5 $\%$ from the MVO.
The inlier ratio of the MRSVO is about 10 $\%$ larger than those of the MVO because the MRSVO suppresses the noise as well as RS distortion.
Figure \ref{fig:real_history_inlier} shows that the number of inliers in test sequences.
The MRSVO provides larger numbers of inliers compared to the MVO overall. 
Figure \ref{fig:real_qualitative_inlier_comparison} compares the inliers of the MRSVO and the MVO on the images in sequence 21.
Inlier ratios annotated on images clearly demonstrate the superiority of the MRSVO.

\section{Conclusion} \label{sec:conclusion}

The MVO with an RS camera suffers from the undesirable artifacts when the camera moves quickly.
To resolve this problem, we proposed a novel MVO algorithm considering the geometric characteristics of RS camera. 
The main idea is the joint estimation of the relative transformation and instantaneous camera motion in two consecutive images.
The proposed algorithm provides accurate ego-motion in an online manner in the presence of severe RS distortions.
The superiority of the proposed algorithm has been verified through expensive experiments on synthetic and real datasets. 
The results of the proposed algorithm can be utilized as an initial state of offline/time-delayed ego-motion estimation algorithms.
%

%\addtolength{\textheight}{-12cm} 

\clearpage

\bibliographystyle{splncs}
\bibliography{mrsvo}
\end{document}